\documentclass[10pt,twocolumn,letterpaper]{article}

\usepackage[pagenumbers]{iccv} 

\definecolor{iccvblue}{rgb}{0.21,0.49,0.74}
\usepackage[pagebackref,breaklinks,colorlinks,allcolors=iccvblue]{hyperref}

\usepackage{cuted}
\usepackage{multirow}
\usepackage{amssymb}
\usepackage{float}
\usepackage{xcolor}
\definecolor{softyellow}{RGB}{218,165,32}

\def\paperID{7872} 
\def\confName{ICCV}
\def\confYear{2025}

\title{DiffVSR: Revealing an Effective Recipe for Taming Robust Video Super-Resolution Against Complex Degradations}

\author{
    Xiaohui Li$^{1,2*}$\quad
    Yihao Liu$^{2*\textsuperscript{$\dagger$}}$\quad
    Shuo Cao$^{3,2}$\quad
    Ziyan Chen$^{4}$\quad
    Shaobin Zhuang$^{1}$\quad\\
    Xiangyu Chen$^{2}$\quad
    Yinan He$^{2}$\quad
    Yi Wang$^{2}$\quad
    Yu Qiao$^{2,4}$\quad
    \\
    \footnotesize{$^1$Shanghai Jiao Tong University
    \quad $^2$Shanghai Artificial Intelligence Laboratory
    \quad} 
    \footnotesize{$^3$University of Science and Technology of China \quad} \\
    \footnotesize{$^4$Shenzhen Institute of Advanced Technology, Chinese Academy of Sciences \quad} \\
    {\tt\small \url{https://xh9998.github.io/DiffVSR-project/}}
}

\begin{document}
\maketitle

\begin{strip}
\vspace{-1.9cm}
\centering
\includegraphics[width=\textwidth]{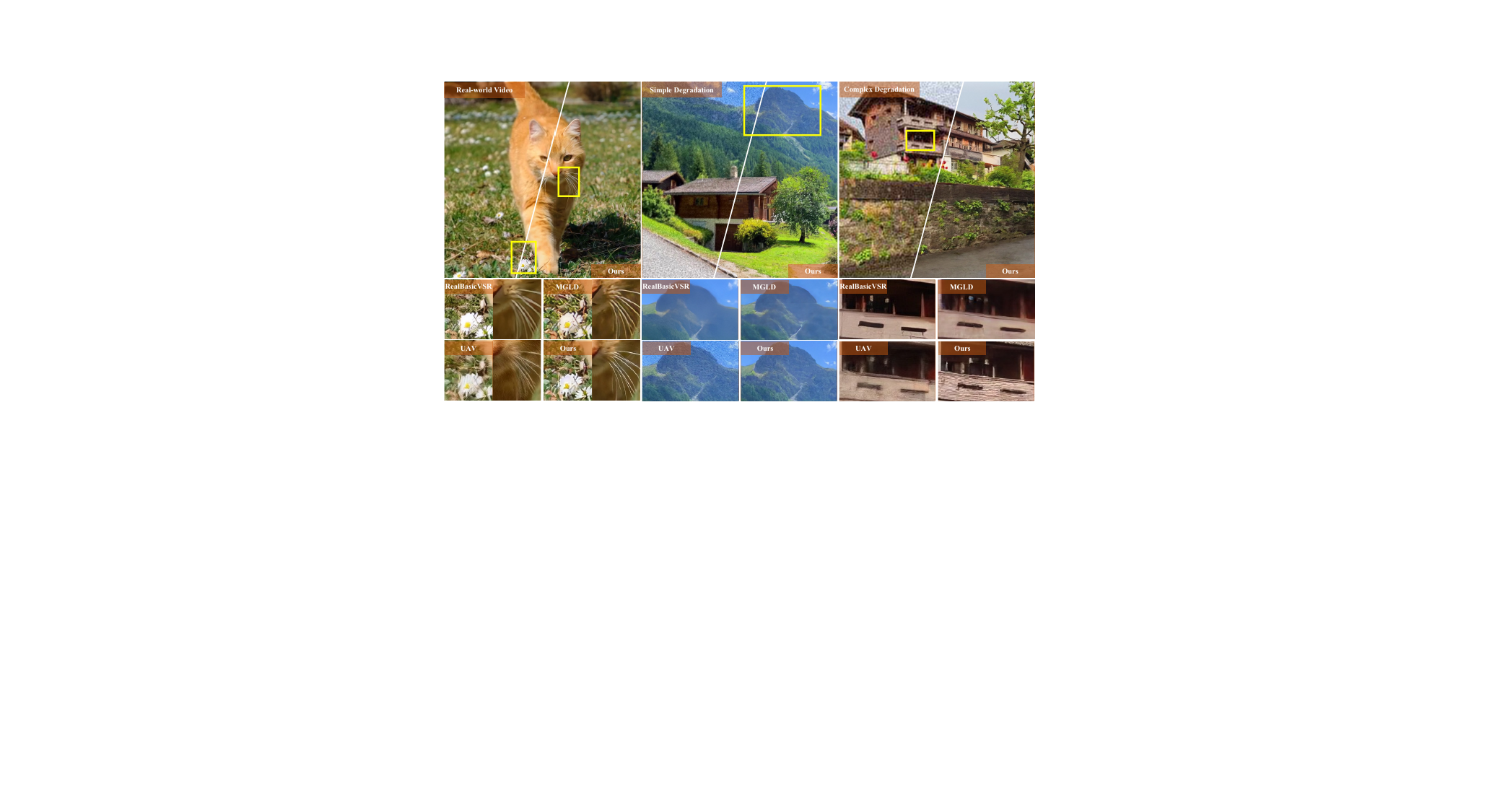}
\captionof{figure}{\textbf{Motivation: Limitations of Existing VSR Methods on Complex Degradations.} As degradation complexity increases, state-of-the-art methods demonstrate significant performance drop—either producing over-smoothed results (oil painting effect) or failing to remove complex artifacts. This limitation persists across different architectural designs, revealing that architectural innovation alone is insufficient for handling complex real-world degradations. Our work addresses this fundamental challenge. \textbf{(Zoom in for best view)}}
\label{fig:teaser}
\end{strip}

\def\thefootnote{*}\footnotetext{Equal contribution. \textsuperscript{$\dagger$}Corresponding author.}

\begin{abstract}
Diffusion models have demonstrated exceptional capabilities in image restoration, yet their application to video super-resolution (VSR) faces significant challenges in balancing fidelity with temporal consistency. Our evaluation reveals a critical gap: existing approaches consistently fail on severely degraded videos--precisely where diffusion models' generative capabilities are most needed. We identify that existing diffusion-based VSR methods struggle primarily because they face an overwhelming learning burden: simultaneously modeling complex degradation distributions, content representations, and temporal relationships with limited high-quality training data. To address this fundamental challenge, we present DiffVSR, featuring a Progressive Learning Strategy (PLS) that systematically decomposes this learning burden through staged training, enabling superior performance on complex degradations. Our framework additionally incorporates an Interweaved Latent Transition (ILT) technique that maintains competitive temporal consistency without additional training overhead. Experiments demonstrate that our approach excels in scenarios where competing methods struggle, particularly on severely degraded videos. Our work reveals that addressing the learning strategy, rather than focusing solely on architectural complexity, is the critical path toward robust real-world video super-resolution with diffusion models.
\end{abstract}

\section{Introduction}

The rapid growth of digital video content across media platforms has heightened the demand for high-quality video viewing experiences. Real-world video super-resolution (VSR), which aims to reconstruct high-resolution (HR) videos from low-resolution (LR) ones with complex degradations, has become increasingly important in both academic research and industrial applications. Unlike the controlled environments of synthetic datasets, real-world videos exhibit diverse and compound degradations--including noise, compression artifacts, blur, and their various combinations--presenting significant challenges for existing VSR methods.

Recent advancements in diffusion models have demonstrated exceptional capabilities in image generation and restoration tasks \cite{rombach2022high, saharia2022photorealistic}, suggesting their potential for VSR applications. Despite this promise, a critical gap exists between claimed and actual performance: while numerous VSR methods report effectiveness on real-world degradations in their controlled experiments, our evaluations reveal consistent failure when confronted with severely degraded real-world videos. These methods either produce over-smoothed results resembling oil paintings or fail to remove complex artifacts--precisely in scenarios where diffusion models' generative power is most needed.

Our investigations reveal a significant insight: despite the diversity of architectural approaches in recent works (e.g., MGLD, UAV with different backbones), they all struggle with complex real-world degradations. This suggests that the primary bottleneck may not lie in architectural design. Instead, we identify the fundamental challenge as the overwhelming learning burden placed on diffusion models, which must simultaneously learn complex degradation distributions, content representations, temporal relationships, and perceptual quality optimization--all with limited high-quality video training data.

This insight prompted us to reconsider the conventional wisdom in diffusion-based VSR research. Rather than focusing on architectural refinements, we identified that \textbf{how models learn} fundamentally determines performance on complex degradations. Based on this insight, we present DiffVSR, featuring our core innovation: a Progressive Learning Strategy (PLS) that gradually builds the model's capacity to handle increasingly complex degradations by systematically decomposing the learning burden. Additionally, we developed an Interweaved Latent Transition (ILT) approach that ensures seamless integration of video segments without requiring additional training or complex alignment operations.

Our framework also includes architectural components (multi-scale temporal attention and temporal-enhanced VAE) that work synergistically with our learning strategy. Our ablation studies demonstrate that while these architectural elements provide measurable benefits, the gains from PLS are particularly substantial when handling severely degraded videos. This finding explains why many existing approaches with similar or even more complex architectural elements still struggle with severe degradations—they focus on architecture while underestimating the critical importance of how models learn.

Through experimentation on both synthetic and real-world datasets across varying degradation complexities, we demonstrate that DiffVSR achieves robust performance in restoring severely degraded videos. Notably, our approach recovers fine details and textures while maintaining competitive temporal consistency, even for videos with complex degradation patterns where competing methods show significant quality degradation. This validates our hypothesis that properly addressing the learning burden may be more fundamental than architectural sophistication for real-world VSR applications.

The primary contributions of this work are:

\begin{itemize}
    \item \textbf{Progressive Learning Strategy (PLS)}, our core innovation that enables diffusion models to effectively handle complex real-world degradations by gradually building capacity, providing significantly enhanced robustness compared to existing approaches.
    
    \item \textbf{Interweaved Latent Transition (ILT)} that complements PLS by ensuring seamless integration of video segments without additional training or computational overhead, addressing a practical limitation in sequence processing.
    
    \item Extensive evaluations showing our method's superior \textbf{robustness across varying degradation complexities}, establishing a new practical recipe for real-world VSR applications.
\end{itemize}

Our work establishes an effective approach for real-world VSR and highlights the importance of learning strategies in diffusion-based video restoration. We encourage readers to view our \textit{\textbf{supplementary video results}}, which demonstrate our method's ability to recover fine details while maintaining competitive temporal consistency more effectively than static figures.
\begin{figure*}
  \centering
   \includegraphics[width=\linewidth]{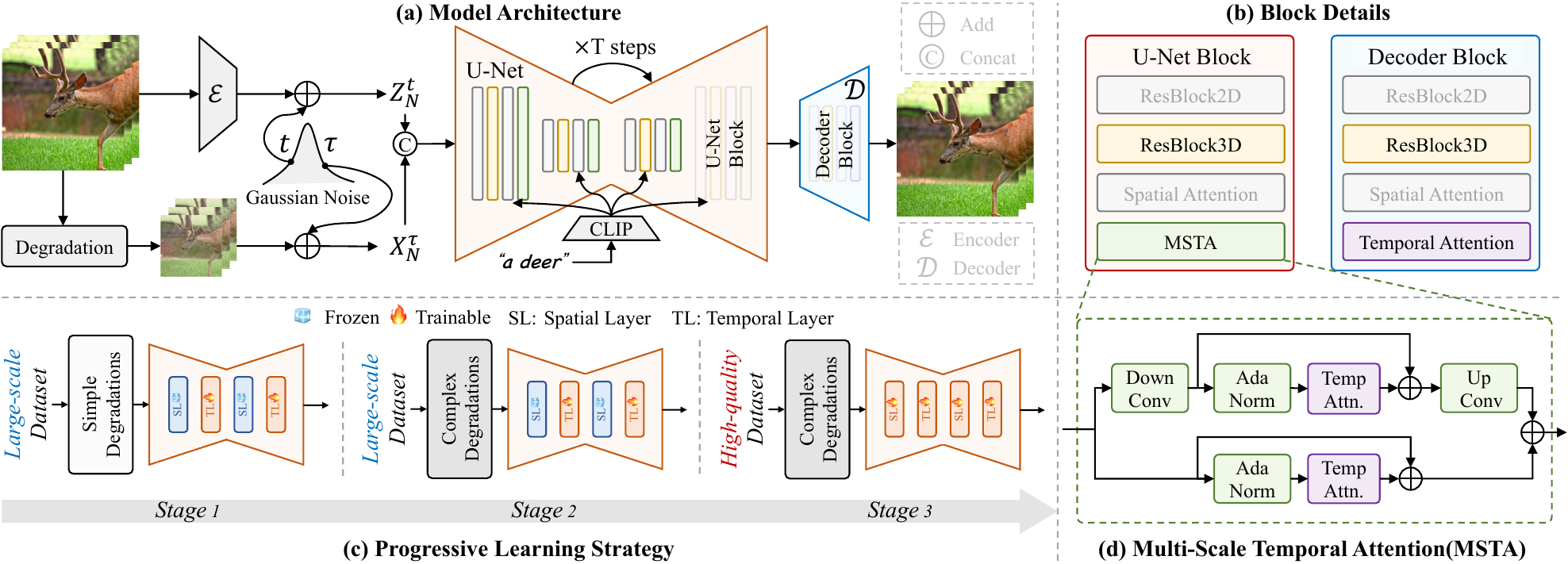}

    \caption{Overview of our proposed DiffVSR framework. (a) Model architecture with enhanced UNet and VAE. (b) Architectural improvements for feature extraction and reconstruction. (c) Progressive Learning Strategy (PLS), our core innovation for handling complex degradations. (d) Multi-Scale Temporal Attention (MSTA) for capturing temporal dependencies at different scales.}
   \label{fig:method}
\end{figure*}

\section{Related Work}
\label{sec:related_work}

\textbf{Video Super-Resolution.} Traditional VSR methods focus on sophisticated architectures to leverage temporal information. TDAN \cite{tian2020tdan} and EDVR \cite{wang2019edvr} introduced deformable convolution for feature alignment, while BasicVSR \cite{chan2021basicvsr} and BasicVSR++ \cite{chan2022basicvsr++} enhanced temporal modeling with recurrent structures. Despite advances in architectural design, these methods often struggle with complex real-world degradations.

\noindent\textbf{Diffusion Models for Image Restoration.} Diffusion models have revolutionized image restoration by providing powerful generative priors. These models \cite{lin2023diffbir, wang2024exploiting, fei2023generative} demonstrate exceptional performance on severely degraded images. However, extending these capabilities to video introduces unique challenges, as the inherent randomness can compromise temporal consistency.

\noindent\textbf{Diffusion Models for Video Restoration.} Recent approaches have attempted to address temporal consistency through architectural innovations. Upscale-A-Video \cite{zhou2024upscale} employs temporal layers and recurrent propagation, while MGLD \cite{yang2025motion} incorporates optical flow guidance. Our evaluation reveals these methods still struggle with severely degraded videos--not primarily due to architectural deficiencies but from the overwhelming learning burden of simultaneously modeling degradations, content, and temporal relationships with limited training data. Our work addresses this [fundamental] challenge by rethinking how diffusion models learn rather than focusing solely on architecture.

\section{Method}

Despite the proliferation of architecturally diverse video super-resolution (VSR) methods, existing approaches consistently struggle with severely degraded videos. This persistent gap motivated our investigation beyond architectural sophistication, leading to a key discovery: the primary bottleneck in diffusion-based VSR lies in the overwhelming learning burden—simultaneously handling complex degradation patterns, content representations, and temporal relationships with limited high-quality training data.

Our DiffVSR framework, shown in Fig.~\ref{fig:method}, introduces a Progressive Learning Strategy (PLS) that systematically decomposes the learning burden across degradation complexity, dataset quality, and parameter optimization. This core innovation enables gradual capacity building, significantly improving performance on severely degraded inputs. As a complement, our Interweaved Latent Transition (ILT) approach efficiently ensures temporal consistency across video segments without additional training overhead. Experiments demonstrate that PLS, rather than architectural sophistication, is the dominant factor unlocking diffusion models' latent capabilities for complex restoration tasks—challenging conventional wisdom in video restoration research.

\subsection{Preliminary: Generative Diffusion Prior}
\label{subsec:sdup}

Following pretrained prior discussions in~\cite{zhou2024upscale}, we select Stable Diffusion x4 Upscaler as our backbone to showcase PLS's significant impact. Our model builds upon this pretrained large-scale text-to-image latent diffusion model (LDM)~\cite{rombach2022stable}, which employs an autoencoder converting images $x$ into low-dimensional latents $z$ with an encoder $\mathcal{E}$ and reconstructing them with a decoder $\mathcal{D}$. A conditional denoising U-Net forms the core, operating in compressed latent space.

During training, for latent samples $z \sim p_{data}$, Gaussian noise follows a predefined schedule to generate noisy latents $z_t = \alpha_t z + \sigma_t \epsilon$, where $\epsilon \sim \mathcal{N}(0, I)$, and $\alpha_t$, $\sigma_t$ define the noise schedule at timestep $t$. Low-resolution inputs $x$ are also noise-perturbed as $x_\tau = \alpha_\tau x + \sigma_\tau \epsilon$ (where $\tau$ corresponds to early diffusion steps) to enhance detail generation. Using v-prediction parameterization~\cite{salimans2022progressive}, the U-Net denoiser $f_\theta$ is optimized to predict $v_t \equiv \alpha_t\epsilon - \sigma_tx$ by minimizing:

\begin{equation}
\mathcal{L} = \mathbb{E}_{z,x,c,t,\epsilon}\left[\|v_t - f_\theta(z_t, x_\tau; c, t)\|_2^2\right],
\end{equation}

where $c$ represents conditional inputs including text prompts and noise levels. At inference, the model iteratively denoises latent representations conditioned on low-resolution inputs, with flexible control over the sampling process through text prompts and noise scheduling.

\subsection{Progressive Learning Strategy}
\label{subsec:pls}

Our key insight identifies the primary VSR performance bottleneck: the overwhelming learning burden on diffusion models to simultaneously handle degradation distributions, content representations, and temporal relationships. While existing approaches emphasize architectural innovations, our Progressive Learning Strategy (PLS) addresses this critical gap by systematically decomposing the learning process across three dimensions: degradation complexity, dataset quality, and parameter optimization.

\noindent\textbf{Stage 1: Temporal Layer Fine-tuning.} We begin by freezing all spatial layers of the pretrained image diffusion model, focusing exclusively on temporal layer fine-tuning using large-scale data. At this stage, we introduce only simple degradations (Gaussian blur and bicubic downsampling)—allowing the model to establish temporal consistency foundations before tackling severe degradations. This controlled approach to parameter optimization and degradation complexity enables the model to build temporal understanding first, creating a foundation for more complex restoration tasks.

\noindent\textbf{Stage 2: Complex Degradation Adaptation.} Building on the temporal consistency established in Stage 1, we progressively increase degradation complexity by introducing noise, compression artifacts, and other real-world distortions~\cite{chan2022investigating,wang2021real} while maintaining dataset scale. This staged introduction prevents the model from being overwhelmed by simultaneous learning of temporal relationships and complex degradation patterns, facilitating more effective adaptation to challenging real-world conditions.

\noindent\textbf{Stage 3: High-quality Refinement.} In the final stage, we elevate both dataset quality and optimization strategy by fully unfreezing all parameters and fine-tuning the entire model with high-quality video data containing complex degradations. This comprehensive refinement across all three dimensions enables precise adaptation to real-world conditions, generating sharper results with refined details and reduced artifacts.

Our ablation studies (Sec.~\ref{sec:ablation}) confirm that this progressive approach significantly outperforms direct training on complex degradations, validating that properly addressing the model's learning burden is more critical than architectural sophistication for handling severely degraded videos.

\begin{figure}
  \centering
   \includegraphics[width=\linewidth]{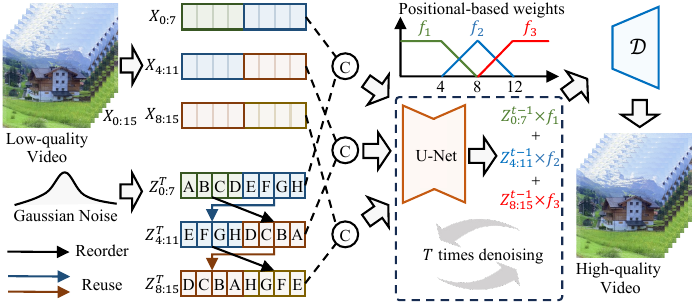}
   \caption{Interweaved Latent Transition approach illustrated. By combining strategic noise rescheduling across overlapping regions with position-based latent interpolation between adjacent subsequences, this lightweight solution ensures temporal consistency without requiring additional training or computational resources.}
   \label{fig:ILT}
\end{figure}

\subsection{Interweaved Latent Transition}
\label{subsec:ilt}

To ensure seamless temporal consistency across video segments while complementing our PLS approach, we introduce the Interweaved Latent Transition (ILT) technique. This addresses a fundamental challenge in processing long videos: the need to segment sequences often creates boundary inconsistencies that compromise visual quality.

As Fig.~\ref{fig:ILT} shows, we divide each video into overlapping subsequences $\{L_0, L_1, \ldots, L_N\}$. Processing each subsequence $L_i$ through the UNet yields corresponding latent feature sequences $F_i$.

For consecutive latent sequences $F_i$ and $F_{i+1}$, our ILT performs position-based interpolation within overlapping frames. For the $j$-th frame in the overlap region, we compute the fused latent as:

\begin{equation}
    F_{fused}[j] = \alpha_j F_{i}[s+j] + (1-\alpha_j) F_{i+1}[j],
\end{equation}

where $\alpha_j= 1-\frac{j}{(P-1)}$ represents position-based weighting, $P$ is the overlap length, and $s$ is the stride between windows. Our implementation uses $N_{test} = 8$ frames per subsequence with 4-frame overlaps, balancing smooth transitions with computational efficiency.

\noindent\textbf{Noise rescheduling.} To further strengthen temporal coherence, we integrate a noise rescheduling mechanism in ILT. Research~\cite{ho2022video,wu2023tune,qiu2023freenoise} shows that temporal consistency in video diffusion models depends significantly on both input content and initial sampling noise. We initialize $N_{test}$ random noise frames $[\epsilon_1, \epsilon_2, \ldots, \epsilon_{N_{test}}]$ for the first sequence, then strategically reuse and reorder these for subsequent subsequences' overlapping regions. This synchronizes the diffusion process across frames, minimizing temporal jitter without additional model training or computational cost.

\subsection{Architectural Components}
\label{subsec:arch}

While our Progressive Learning Strategy represents our primary contribution, our framework incorporates supportive architectural components that enhance video restoration effectiveness. These elements work synergistically with PLS to boost overall performance.

\noindent\textbf{Multi-Scale Temporal Attention.} To achieve balance between temporal consistency and detailed texture generation, we implement a multi-scale temporal attention (MSTA) module. As shown in Fig.~\ref{fig:method}(d), this module fuses information across multiple scales to capture diverse motion patterns and textural details.

For an intermediate U-Net hidden embedding $h$, we generate a downsampled version $h_{\downarrow} = \texttt{DConv}(h)$ using strided convolution. We then apply temporal attention separately to both resolutions:

\begin{equation}
\begin{aligned}
h_{down} &=\texttt{TA}(\texttt{AdaLN}(h_{\downarrow}))+h_{\downarrow}, \\
h_{ori} &=\texttt{TA}(\texttt{AdaLN}(h))+h,
\end{aligned}
\end{equation}

where $\texttt{TA}$ denotes temporal attention operations and $\texttt{AdaLN}$ represents adaptive layer normalization. The final hidden embedding $h'$ combines the original temporal attention output with the upsampled downscale branch:

\begin{equation}
    h'=h_{down} +  \alpha \cdot \texttt{UPConv}(h_{down}),
\end{equation}

with $\texttt{UPConv}$ handling upsampling and $\alpha$ (set to 0.5) controlling fusion weight.

\noindent\textbf{Temporal-Enhanced VAE.} To strengthen temporal consistency in the decoder, we developed a Temporal-Enhanced 3DVAE (TE-3DVAE) that extends the original 2D VAE architecture with 3D residual blocks and temporal attention layers. This enhancement captures both short-term and long-term temporal dependencies more effectively. We train these VAE variants using a combination of L1 reconstruction loss, perceptual loss~\cite{zhang2018unreasonable}, and adversarial loss with a temporal PatchGAN discriminator~\cite{isola2017image}.

Our ablation studies (Tab.~\ref{tab:ablation}) reveal a critical insight: while architectural improvements yield measurable benefits, PLS delivers substantially larger performance gains by unlocking the model's latent capabilities, especially for severely degraded videos. This finding highlights a current research imbalance where architectural sophistication often receives disproportionate focus compared to learning strategies. Our work demonstrates the synergistic relationship between these aspects—optimized learning approaches enable architectural innovations to achieve their full potential.

\begin{figure}[ht]
  \centering
   \includegraphics[width=0.95\columnwidth]{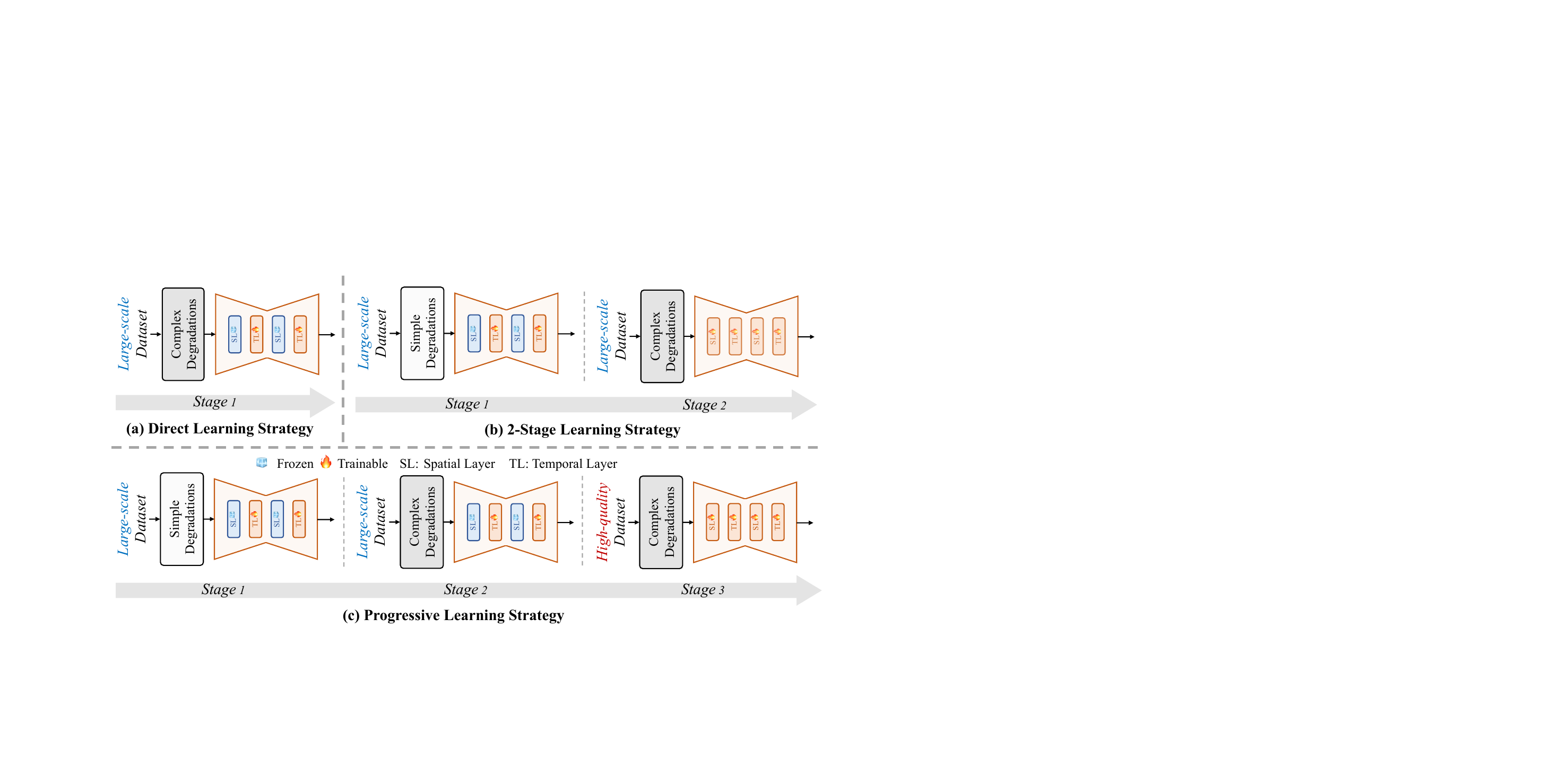}

\caption{Illustration of three training strategy variants: direct training, partial progressive learning, and our full PLS approach.}
   \label{fig:pls_abl}
\end{figure}

\begin{table}[ht]
\centering
\resizebox{\columnwidth}{!}{%
\begin{tabular}{c|ccccc}
\toprule
Exp. & PSNR↑ & CLIP-IQA↑ & MUSIQ↑ & DOVER↑ & $E_{warp}^*$↓ \\
\hline
(1) & 23.299 & 0.8432 & 68.182 & 0.864 & 2.582 \\
(2) & 23.591 & 0.8383 & 68.449 & 0.866 & 2.236 \\
(3) & \textbf{24.172} & \textbf{0.8666} & \textbf{68.982} & \textbf{0.870} & \textbf{1.707} \\
\bottomrule
\end{tabular}
}
\caption{Comparison of training strategies: direct training (1), partial progressive learning (2), and full PLS (3). Our complete PLS approach achieves the best results across all image quality and temporal consistency metrics.}
\label{tab:pls}
\end{table}

\section{Experiments}
\label{Experiments}

\begin{table*}[ht]
\centering
\resizebox{\textwidth}{!}{%
\begin{tabular}{c|c|ccccccc}
\toprule
Datasets & Metrics & Real-ESRGAN\cite{wang2021real} & SD ×4 Upscaler\cite{rombach2022stable} & DiffBIR\cite{lin2023diffbir} & RealBasicVSR\cite{chan2022investigating} & MGLD\cite{yang2025motion} & UAV\cite{zhou2024upscale} & Ours \\
\hline
\multirow{5}{*}{\centering UDM10} & PSNR↑ & 26.872 & 24.039 & 26.767 & \textcolor{red}{28.526} & \textcolor{blue}{27.801} & 27.412 & 26.966 \\
& CLIP-IQA↑ & 0.708 & \textcolor{red}{0.775} & 0.679 & 0.735 & 0.695 & 0.681 & \textcolor{blue}{0.736} \\
& MUSIQ↑ & 61.792 & \textcolor{blue}{64.689} & 64.562 & 64.248 & 60.470 & 56.633 & \textcolor{red}{67.731} \\
& DOVER↑ & 0.815 & \textcolor{red}{0.827} & 0.784 & 0.814 & 0.780 & 0.723 & \textcolor{blue}{0.826} \\
& $E_{warp}^* \downarrow$ & 1.139 & 3.173 & 1.793 & \textcolor{red}{0.499} & 1.068 & 0.661 & \textcolor{blue}{0.604} \\
\hline
\multirow{5}{*}{\centering \shortstack{UDM10 \\ (complex deg.)}} & PSNR↑ & 22.550 & 22.299 & 22.781 & \textcolor{red}{23.380} & \textcolor{blue}{23.342} & 22.053 & 23.181 \\
& CLIP-IQA↑ & 0.557 & 0.477 & 0.475 & 0.570 & 0.613 & \textcolor{blue}{0.663} & \textcolor{red}{0.726} \\
& MUSIQ↑ & 55.801 & 32.417 & 59.290 & \textcolor{blue}{60.623} & 58.481 & 57.117 & \textcolor{red}{66.431} \\
& DOVER↑ & 0.699 & 0.294 & 0.616 & \textcolor{blue}{0.719} & 0.657 & 0.675 & \textcolor{red}{0.826} \\
& $E_{warp}^* \downarrow$ & 2.74 & 3.93 & 3.887 & \textcolor{red}{1.232} & 2.301 & 2.259 & \textcolor{blue}{1.384} \\
\hline
\multirow{5}{*}{\centering YouHQ40} & PSNR↑ & 24.421 & 21.687 & 24.154 & \textcolor{red}{24.818} & \textcolor{blue}{24.524} & 23.529 & 23.713 \\
& CLIP-IQA↑ & 0.766 & 0.752 & 0.722 & \textcolor{red}{0.797} & 0.745 & 0.708 & \textcolor{blue}{0.785} \\
& MUSIQ↑ & 62.050 & 62.623 & \textcolor{blue}{67.522} & 65.398 & 63.407 & 57.141 & \textcolor{red}{68.040} \\
& DOVER↑ & 0.876 & 0.854 & 0.840 & \textcolor{red}{0.885} & 0.847 & 0.815 & \textcolor{blue}{0.878} \\
& $E_{warp}^* \downarrow$ & 2.167 & 5.832 & 2.765 & \textcolor{red}{1.065} & \textcolor{blue}{1.362} & 1.615 & 1.492 \\
\hline
\multirow{5}{*}{\centering \shortstack{YouHQ40 \\ (complex deg.)}} & PSNR↑ & 22.122 & 21.034 & \textcolor{blue}{22.347} & 22.247 & \textcolor{red}{22.503} & 21.836 & 22.159 \\
& CLIP-IQA↑ & 0.715 & 0.718 & 0.651 & \textcolor{blue}{0.745} & 0.724 & 0.659 & \textcolor{red}{0.767} \\
& MUSIQ↑ & 58.866 & 47.475 & 65.157 & \textcolor{blue}{66.471} & 57.219 & 57.469 & \textcolor{red}{67.679} \\
& DOVER↑ & 0.634 & 0.092 & 0.624 & \textcolor{blue}{0.676} & 0.594 & 0.606 & \textcolor{red}{0.710} \\
& $E_{warp}^* \downarrow$ & 3.388 & 4.772 & 4.000 & 1.739 & 1.753 & \textcolor{blue}{1.668} & \textcolor{red}{1.634} \\
\hline
\multirow{4}{*}{\centering MVSR4x} & NRQM↑ & 4.731 & 3.081 & \textcolor{red}{6.564} & 5.746 & 4.665 & 5.582 & \textcolor{blue}{6.432} \\
& CLIP-IQA↑ & 0.460 & 0.397 & 0.407 & 0.426 & \textcolor{blue}{0.487} & 0.395 & \textcolor{red}{0.491} \\
& MUSIQ↑ & 54.331 & 22.670 & 62.599 & \textcolor{blue}{62.716} & 50.367 & 56.128 & \textcolor{red}{66.829} \\
& DOVER↑ & 0.634 & 0.092 & 0.624 & \textcolor{blue}{0.676} & 0.594 & 0.606 & \textcolor{red}{0.710} \\
\hline
\multirow{4}{*}{\centering RealVideo10} & NRQM↑ & 4.495 & 4.801 & \textcolor{red}{6.880} & 5.680 & 4.768 & 5.399 & \textcolor{blue}{5.805} \\
& CLIP-IQA↑ & 0.831 & 0.846 & 0.803 & 0.836 & \textcolor{red}{0.861} & 0.835 & \textcolor{blue}{0.856} \\
& MUSIQ↑ & 49.733 & 39.920 & 58.647 & \textcolor{red}{61.121} & 52.332 & 50.029 & \textcolor{blue}{59.566} \\
& DOVER↑ & 0.704 & 0.485 & 0.716 & \textcolor{red}{0.773} & 0.714 & 0.610 & \textcolor{blue}{0.749} \\
\bottomrule
\end{tabular}%
}
\caption{Quantitative comparison with state-of-the-art methods on synthetic datasets (UDM10 and YouHQ40, each with simple and complex degradation) and real-world datasets (MVSR4x, RealVideo10). The best and second-best results are marked in \textcolor{red}{red} and \textcolor{blue}{blue}.}
\label{tab:quality}
\end{table*}

\begin{figure*}[ht]
  \centering
   \includegraphics[width=\linewidth]{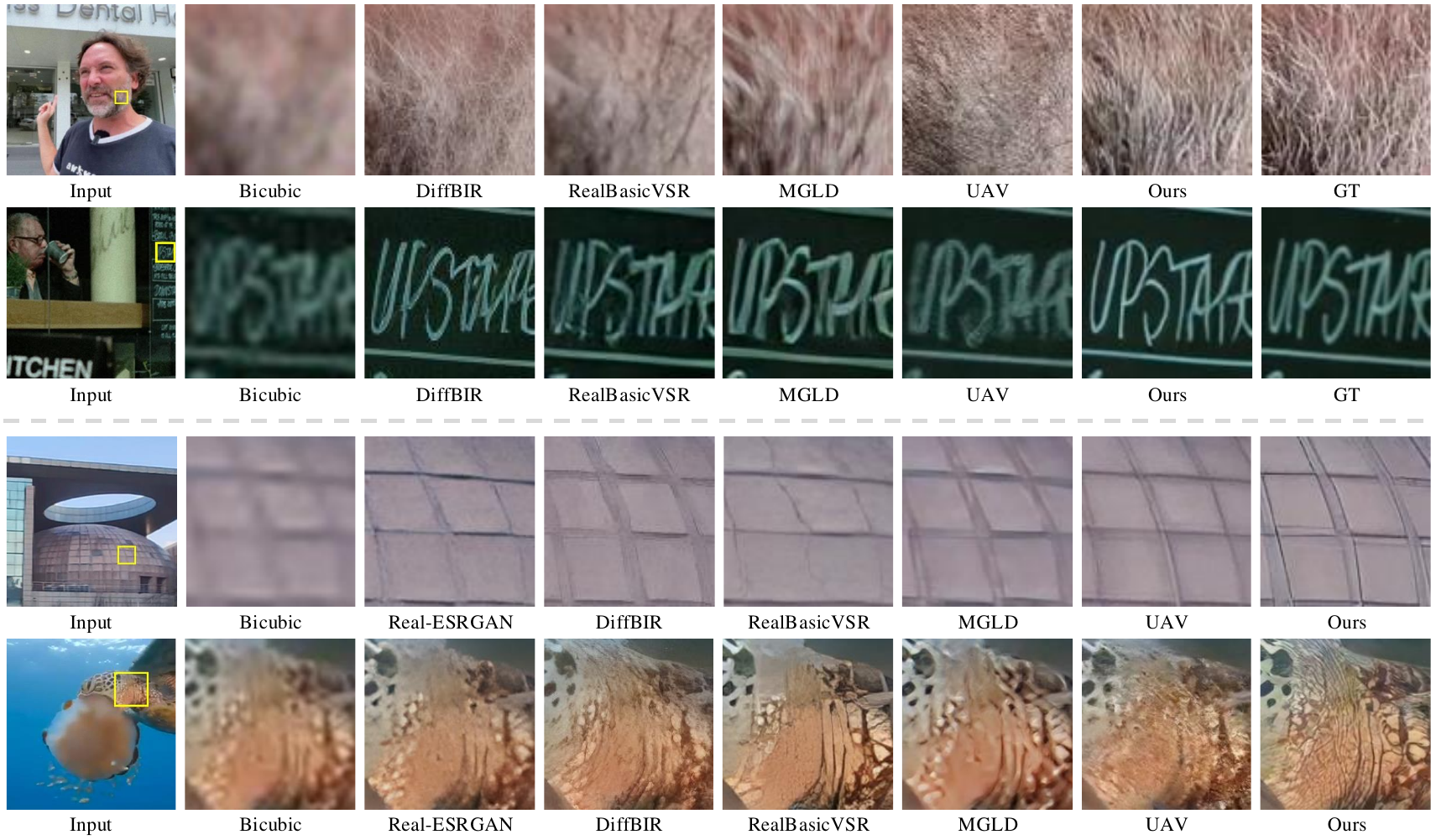}

    \caption{Visual comparison on synthetic (top) and real-world (bottom) degraded videos. Our method achieves clear and natural restoration of text, building texture details, human hair, and animal fur \textbf{under severe degradation}, while competing methods either fail to remove artifacts or generate unnatural oil-painting-like details. \textbf{(Zoom-in for best view)}}
   \label{fig:syscomp}
\end{figure*}

\subsection{Datasets and Implementation Details}

\noindent\textbf{Datasets.} For stages 1 and 2 training, we utilize a WebVid2M~\cite{bain2021frozen} subset containing approximately 400K video-text pairs at 336×596 resolution. Stage 3 fine-tuning employs OpenVid-1M~\cite{nan2024openvid} with 1M high-resolution (512×512 or larger) text-video pairs and YouHQ~\cite{zhou2024upscale} featuring 37K 2K-resolution videos without text annotations. Low-quality inputs are generated using RealBasicVSR~\cite{chan2022investigating} degradation pipeline. We evaluate on both synthetic datasets (UDM10~\cite{PFNL}, YouHQ40~\cite{zhou2024upscale} with varied degradation levels) and real-world collections (MVSR4x~\cite{wang2023benchmark} and our RealVideo10 dataset featuring diverse scenes).

\noindent\textbf{Training Details.} Our PyTorch-based model is trained on 8 NVIDIA A100 GPUs using AdamW~\cite{kingma2014adam} optimizer with 1e-4 learning rate and 96 batch size. We generate LR-HR training pairs by randomly cropping 320×320 patches from 8-frame video segments, with temporal sampling strides varying from 1 to 6 to capture diverse motion patterns. Our model performs 4× video super-resolution.

\noindent\textbf{Evaluation Metrics.} We assess quality using fidelity-based (PSNR) and perceptual-driven metrics (CLIP-IQA~\cite{wang2023exploring}, MUSIQ~\cite{ke2021musiq}, NRQM~\cite{ma2017learning}, and DOVER~\cite{wu2023exploring}). Temporal consistency is evaluated using warping error ($E_{warp}^*$)~\cite{lai2018learning}.

\subsection{Ablation Study} 
\label{sec:ablation}

\begin{figure}[ht]
  \centering
   \includegraphics[width=0.95\columnwidth]{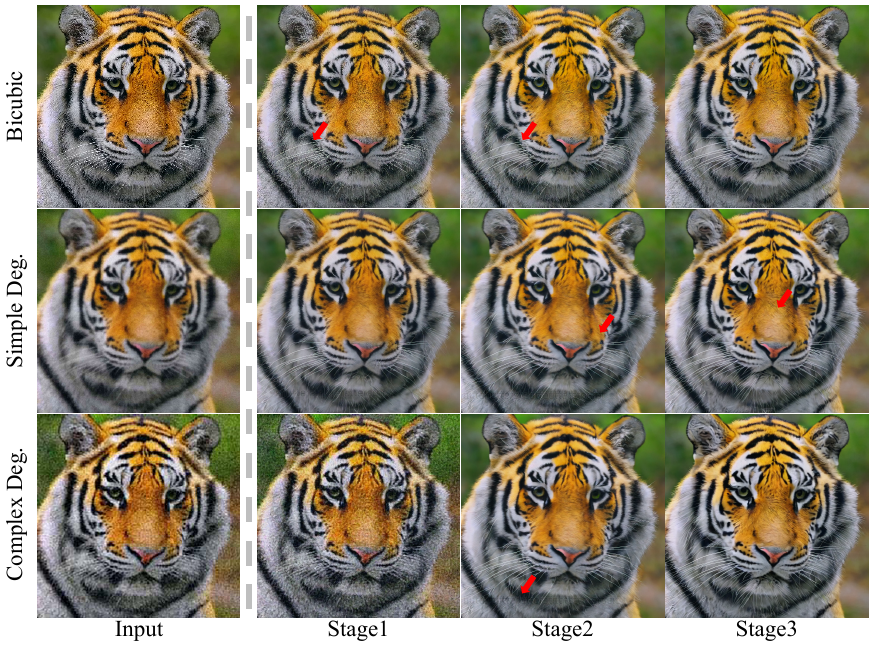}

    \caption{Visual comparison of degraded inputs (left of dashed line) and model outputs across PLS stages (right) with increasing degradation complexity (bicubic, simple mixed, and complex mixed). Each stage shows progressive improvement in handling artifacts, details, and content fidelity.}
    \label{fig:pls_degradation}
\end{figure}

\noindent \textbf{Necessity of Each Part in PLS.} To validate the indispensability of each component in our Progressive Learning Strategy, we conducted ablation studies comparing three approaches, as illustrated in Fig.~\ref{fig:pls_abl}: \textit{Direct Learning Strategy:} Training the model on all degradation levels simultaneously with only temporal layers tuned--a common approach in previous works \cite{zhou2024upscale,yang2025motion} attempting to handle full complexity at once. \textit{2-Stage Learning Strategy:} Initially training on simple degradations with only temporal layers tuned, then continuing with complex degradations while unlocking all parameters--addressing degradation complexity and parameter optimization but lacking systematic progression. \textit{Our Progressive Learning Strategy:} A three-stage approach systematically decomposing the learning burden across degradation complexity, dataset quality, and parameter optimization dimensions.

Evaluating these strategies on YouHQ40 dataset videos (Tab.~\ref{tab:pls}), our full Progressive Learning Strategy significantly outperforms alternatives across all metrics. The Direct Learning Strategy struggles with simultaneously handling multiple complex tasks, resulting in inferior restoration quality and temporal inconsistency. The 2-Stage approach improves but falls short with severe degradations and detail preservation. These results confirm our key insight: the primary bottleneck in diffusion-based VSR is not architectural limitations but the overwhelming learning burden placed on models. By decomposing this burden progressively, each component contributes critically to unlocking the diffusion model's potential for robust performance, particularly with severely degraded real-world videos.

\noindent\textbf{Unlocking Diffusion Potential through PLS.} To validate our strategy's effectiveness for visual quality enhancement, we analyzed performance across training stages with varying degradation levels (Fig.~\ref{fig:pls_degradation}).

Our analysis reveals a clear progression in restoration capabilities across stages. The Stage 1 model demonstrates limited effectiveness, preserving artifacts in bicubic inputs and failing with noise despite exposure to various degradation types. The Stage 2 model exhibits substantial improvement by leveraging the foundation established earlier, yet still displays overfitting to WebVideo data characteristics. In contrast, the Stage 3 model achieves superior performance across all metrics by building cumulatively on previous capabilities with fully unlocked parameters and high-resolution training data. This progression illustrates the fundamental challenge conventional approaches face when attempting simultaneous learning of multiple complex tasks.

Comparing experiments (a) and (b) in Tab.~\ref{tab:ablation} further validates PLS's effectiveness, showing simultaneous improvements in fidelity (PSNR), temporal consistency (warping error), and perceptual quality (MUSIQ). These results demonstrate that progressive capability accumulation is essential for unlocking diffusion models' potential in video super-resolution. This sequential decomposition of learning burden leads to substantially better outcomes than attempting to optimize all aspects simultaneously, which has been the primary bottleneck limiting performance in existing VSR approaches regardless of architectural sophistication.

\noindent\textbf{Interweaved Latent Transition.}
Our experiments demonstrate that PLS significantly enhances both visual quality and intra-segment temporal consistency. ILT further addresses inter-frame continuity--a critical challenge in diffusion-based video restoration. Unlike pure generation tasks, VSR benefits from reference information in LR inputs. Our ILT approach leverages this constraint to achieve robust temporal consistency while handling complex degradations.

\begin{figure}[ht]
  \centering
   \includegraphics[width=0.95\columnwidth]{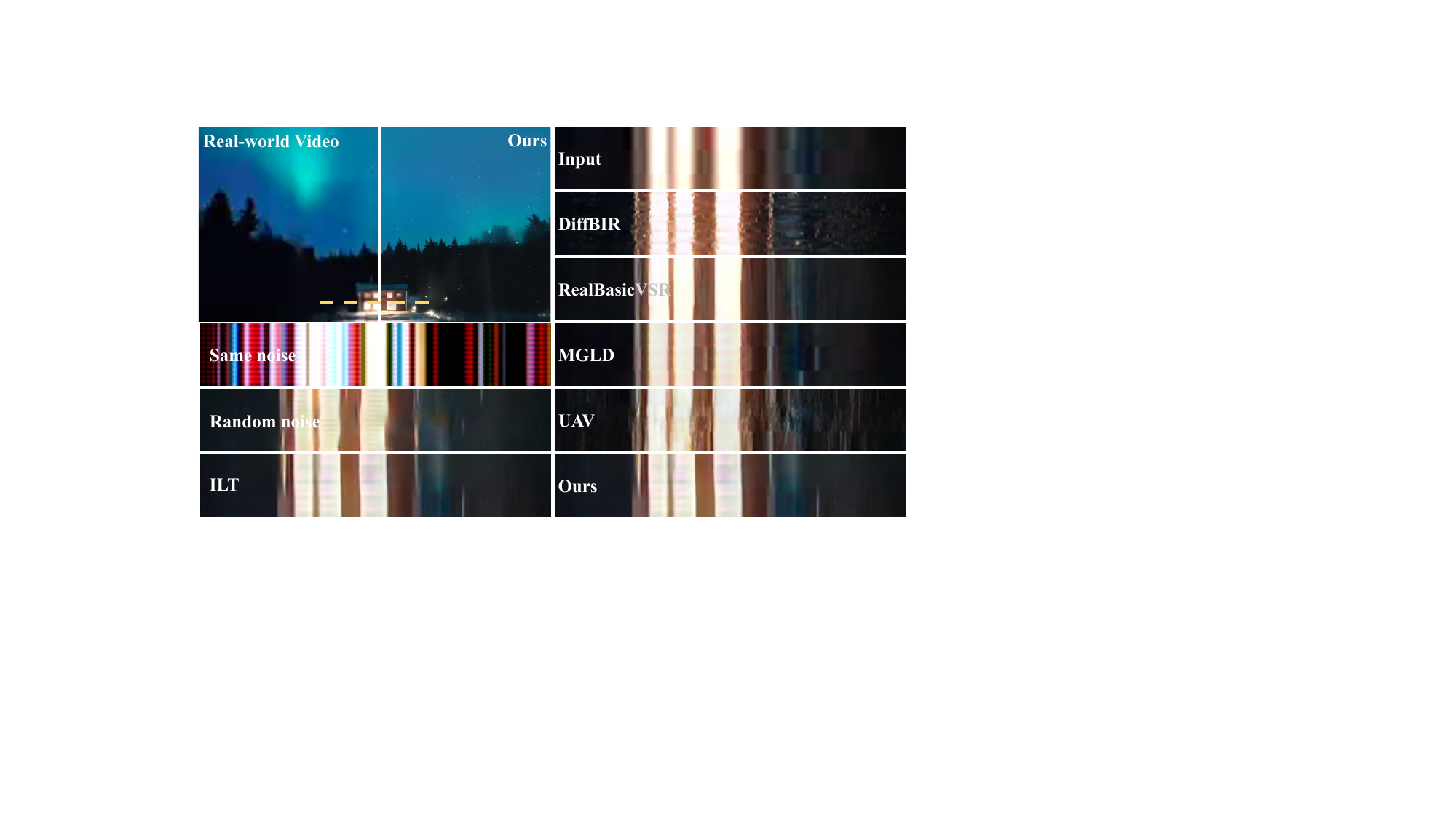}

    \caption{Temporal profiles (stacking the  \textcolor{softyellow}{yellow line} across consecutive frames) demonstrate ILT's advantage over alternative noise transition methods (left) and our full method's superior restoration quality and temporal consistency compared to other approaches (right). \textbf{(Best viewed digitally)}}
    \label{fig:ILT_abl}
\end{figure}

As illustrated in Fig.~\ref{fig:ILT_abl} (right), image-based models exhibit severe temporal discontinuities, while alternative video methods demonstrate blurriness, misalignment, and consistency artifacts. For noise transition, applying identical noise across frames produces catastrophic results (Fig.~\ref{fig:ILT_abl} left, ``Same"). While FreeNoise's \cite{qiu2023freenoise} rearrangement strategy (Fig.~\ref{fig:ILT_abl} left, ``Rearrange") improves continuity, it fails to address input correspondence by neglecting VSR's inherent reference constraints.

Quantitatively, experiments (b) and (c) in Tab.~\ref{tab:ablation} validate ILT's effectiveness: both fidelity and temporal consistency metrics improve substantially. Following the established quality-fidelity tradeoff paradigm, perceptual quality exhibits only a marginal decrease while maintaining superior levels compared to alternatives.

\noindent\textbf{Architectural Improvements.} Analyzing experiments (c) and (d) in Tab.~\ref{tab:ablation}, we observe that our architectural enhancements (TE-3DVAE and MSTA) contribute significant improvements across fidelity, temporal consistency, and visual quality metrics. However, compared to PLS's impact (experiments (a) and (b), where warping error decreases by 0.778 and MUSIQ increases markedly), these architectural contributions function more as complementary refinements than transformative elements.

This evidence supports our central thesis: in video super-resolution, research emphasis should shift toward learning strategies like PLS rather than exclusively pursuing architectural innovations. The substantial performance gains derive from addressing fundamental learning challenges, with architectural optimizations providing incremental refinements. This confirms that maximizing a model's potential necessitates both PLS and architectural improvements operating synergistically, rather than relying predominantly on structural design innovations as has been the prevailing paradigm.

\begin{table}[ht]
\centering
\resizebox{\columnwidth}{!}{%
\begin{tabular}{c|ccc|cccc}
\toprule
Exp. & Arch. & PLS & ILT & PSNR↑ & LPIPS↓ & MUSIQ↑ & $E_{warp}$↓ \\
\hline
(a) & \checkmark & & & 23.570 & 0.319 & 68.907 & 2.834 \\
(b) & \checkmark & \checkmark & & 23.939 & 0.292 & \textbf{69.768} & 2.056 \\
(c) & \checkmark & \checkmark & \checkmark & \textbf{24.172} & \textbf{0.285} & 68.982 & \textbf{1.707} \\
(d) & & \checkmark & \checkmark & 23.971 & 0.292 & 68.597 & 2.290 \\
\bottomrule
\end{tabular}
}
\caption{Ablation study on YouHQ10 (subset of YouHQ40) dataset quantifying the contribution of each component. "Arch." represents architectural improvements (TE-3DVAE and MSTA).}
\label{tab:ablation}
\end{table}


\subsection{Comparisons}
For comprehensive evaluation, we compare our method against state-of-the-art approaches, including both image-based restoration methods (Real-ESRGAN~\cite{wang2021real}, SD ×4 Upscaler~\cite{rombach2022stable}, DiffBIR~\cite{lin2023diffbir}) and video restoration methods (RealBasicVSR~\cite{chan2022investigating}, MGLD~\cite{yang2025motion}, UAV~\cite{zhou2024upscale}).

\noindent\textbf{Quantitative Evaluation.} We evaluate our method against state-of-the-art approaches on synthetic datasets (UDM10 and YouHQ40, with both simple and complex degradation variants) and real-world datasets (MVSR4x, RealVideo10) in Tab.~\ref{tab:quality}. Our model demonstrates excellent performance under standard degradation conditions, while excelling with complex degradations where competing methods typically underperform or produce over-smoothed results with inferior perceptual metrics. This consistent superiority under severe degradation conditions establishes our approach as the definitive state-of-the-art for challenging restoration scenarios, highlighting the exceptional robustness achieved through our Progressive Learning Strategy. Additionally, our method maintains superior performance on real-world datasets, further validating its efficacy in practical applications with complex degradation patterns.

\noindent\textbf{Qualitative Evaluation.} Fig.~\ref{fig:syscomp} presents visual comparisons on both synthetic (top) and real-world (bottom) degraded videos. Under severe degradation conditions, our method consistently delivers superior restoration quality across diverse scenarios. For synthetic degradations, our approach accurately reconstructs fine-grained details such as facial hair textures and precisely restores text on billboards, where competing methods struggle to eliminate artifacts or generate natural-looking results. Similarly, in challenging real-world cases, our method preserves sharp geometric structures while maintaining authentic textures--accurately reconstructing distinctive features such as the sea turtle's skin pattern without introducing artificial textures or stylization artifacts observed in other approaches. These visual results validate our method's exceptional capability to handle extreme degradation while preserving natural details, demonstrating the effectiveness of our progressive learning strategy in real-world restoration tasks. More visual results can be found in the supplementary file.


\section{Conclusion}
In this paper, we present DiffVSR, revealing an effective recipe that fundamentally rethinks how diffusion models should approach robust video super-resolution against complex degradations. Rather than pursuing architectural elaboration, our investigation shows that focusing on how models learn is more critical than what they learn with. Our Progressive Learning Strategy systematically decomposes the overwhelming learning burden faced by diffusion models, enabling them to handle increasingly complex degradations--a capability purely architectural innovations have consistently failed to achieve. Our experiments demonstrate remarkable results in severely degraded scenarios where competing methods falter, confirming our initial hypothesis. By combining PLS with our Interweaved Latent Transition technique, we achieve state-of-the-art perceptual quality while maintaining temporal coherence without additional training overhead. Beyond video super-resolution, our work contributes a valuable insight: the critical bottleneck in diffusion-based restoration may not lie in architecture but in learning strategy. This perspective could benefit adjacent fields like video generation and editing. DiffVSR thus represents not just an advanced VSR method, but a recalibration of research priorities for unlocking diffusion models' full potential in real-world video restoration.

{
    \small
    \bibliographystyle{ieeenat_fullname}
    \bibliography{main}
}
\clearpage
\renewcommand\thesection{\Alph{section}}
\onecolumn
\setcounter{section}{0}

%
\begin{center}
	\Large\textbf{{Appendix}}\\
	\vspace{8mm}
\end{center}

\section{More Ablation Studies}

\subsection{Runtime Analysis for ILT and its alternatives}
Here we examine the computational efficiency of our Interweaved Latent Transition method compared to alternatives. It is well known that incorporating temporal consistency inevitably increases inference time, regardless of the method chosen. As shown in Tab.~\ref{tab:ILT_FA}, achieving an acceptable balance between quality and efficiency is crucial.

Our ILT approach increases runtime compared to the baseline without temporal modeling, but delivers substantial improvements in both quality metrics and temporal consistency. Notably, when compared to flow-based alignment methods commonly used in super-resolution, including those in UAV \cite{zhou2024upscale}, our approach offers several advantages despite similar computational costs: ILT requires no additional training, introduces no extra parameters, and avoids the error accumulation problems inherent in flow-based methods.

The results demonstrate that ILT achieves superior temporal consistency (lower warping error) while maintaining competitive or better performance across other metrics. This favorable trade-off between computational cost and quality improvement aligns with our overall framework philosophy: thoughtful design choices that complement our learning strategy can provide significant benefits without excessive computational burden.

\begin{table}[!ht]
\centering
\resizebox{0.5\columnwidth}{!}{%
\begin{tabular}{c|c|ccc}
\toprule
Datasets & Metrics & w/o ILT & ILT → FA & w/ ILT \\
\hline
\multirow{5}{*}{YouHQ40} 
& PSNR↑ & 23.467 & \textcolor{blue}{23.529} & \textcolor{red}{23.713} \\
& LPIPS↓ & 0.293 & \textcolor{blue}{0.287} & \textcolor{red}{0.288} \\
& MUSIQ↑ & \textcolor{red}{69.257} & 66.195 & \textcolor{blue}{68.040} \\
& $E_{warp}^* \downarrow$ & 1.870 & \textcolor{blue}{1.589} & \textcolor{red}{1.492} \\
& Runtime (s) & \textcolor{red}{424.81} & 728.22 & \textcolor{blue}{727.85} \\
\bottomrule
\end{tabular}
}
\caption{Computational cost and performance comparison of ILT. ILT → FA: replacing ILT with flow alignment in UAV. Runtime: processing video 000, calculated with \textbf{official time computation code}.}
\label{tab:ILT_FA}
\end{table}

\subsection{Effectiveness of Temporal-Enhanced 3DVAE.}
While our primary contribution is the Progressive Learning Strategy, architectural components still provide complementary benefits. We evaluate VAE variants trained on OpenVid-1M using our collected VAE-VAL5 dataset for texture and motion assessment. As shown in Tab.~\ref{tab:VAE}, compared to 2D VAE, the 3D VAE improves PSNR by 0.4dB and reduces warping error by 24\%. Our TE-3DVAE further achieves incremental improvements across all metrics, particularly in temporal consistency. These results demonstrate that architectural enhancements, though secondary to learning strategy, contribute meaningful refinements to the overall framework through better temporal information preservation in the latent space.

\begin{table}[ht]
\centering
\resizebox{0.5\columnwidth}{!}{%
\begin{tabular}{c|cccc}
\toprule
 & PSNR↑ & SSIM↑ & $E_{warp}^* \downarrow$ & TF↑ \\
\hline
2DVAE & 30.251 & 0.917 & 1.049 & 0.965 \\
3DVAE & \textcolor{blue}{30.645} & \textcolor{blue}{0.926} & \textcolor{blue}{0.800} & \textcolor{blue}{0.967} \\
TE-3DVAE & \textcolor{red}{30.874} & \textcolor{red}{0.927} & \textcolor{red}{0.761} & \textcolor{red}{0.968} \\
\bottomrule
\end{tabular}
}
\caption{Quantitative comparison of VAE variants on VAE-VAL5 dataset. While 3D architecture provides substantial improvement over 2D, temporal enhancement offers further refinement, consistent with our thesis that architectural improvements provide complementary benefits to our learning strategy.}

\label{tab:VAE}
\end{table}

\subsection{Effectiveness of Multi-Scale Temporal Attention.}
While our Progressive Learning Strategy is the key to handling complex degradations, architectural components like MSTA provide modest but consistent improvements. As shown in Tab.~\ref{tab:MSTA}, when we remove MSTA while keeping our core learning strategy (Exp. 2), we observe performance decreases across all metrics, with temporal consistency (warping error) showing particularly notable decline (36\% increase in error).

\begin{table}[!ht]
\centering
\resizebox{0.7\columnwidth}{!}{%
\begin{tabular}{c|ccc|cccc}
\toprule
Exp. & MSTA & PLS & ILT & PSNR↑ & LPIPS↓ & MUSIQ↑ & $E_{warp}$↓ \\
\hline
(1) & \checkmark & \checkmark & \checkmark & \textcolor{red}{24.172} & \textcolor{red}{0.285} & \textcolor{red}{68.982} & \textcolor{red}{1.707} \\
(2) & & \checkmark & \checkmark & 23.671 & 0.299 & 68.357 & 2.330 \\
\bottomrule
\end{tabular}
}
\caption{Ablation study on MSTA. While maintaining our core Progressive Learning Strategy, removing MSTA causes performance degradation across all metrics, particularly in temporal consistency.}
\label{tab:MSTA}
\end{table}

The multi-scale design in MSTA applies 2× downsampling to features at intermediate stages, enabling efficient processing of motion information at various scales. This architectural enhancement helps capture temporal dependencies when combined with our learning-focused approach. These results align with our central thesis that while learning strategy is the primary factor for robust performance, well-designed architectural components like MSTA can still provide complementary benefits by improving the model's ability to understand motion dynamics and maintain temporal coherence.

\begin{figure}[htbp]
  \centering
   \includegraphics[width=0.8\linewidth]{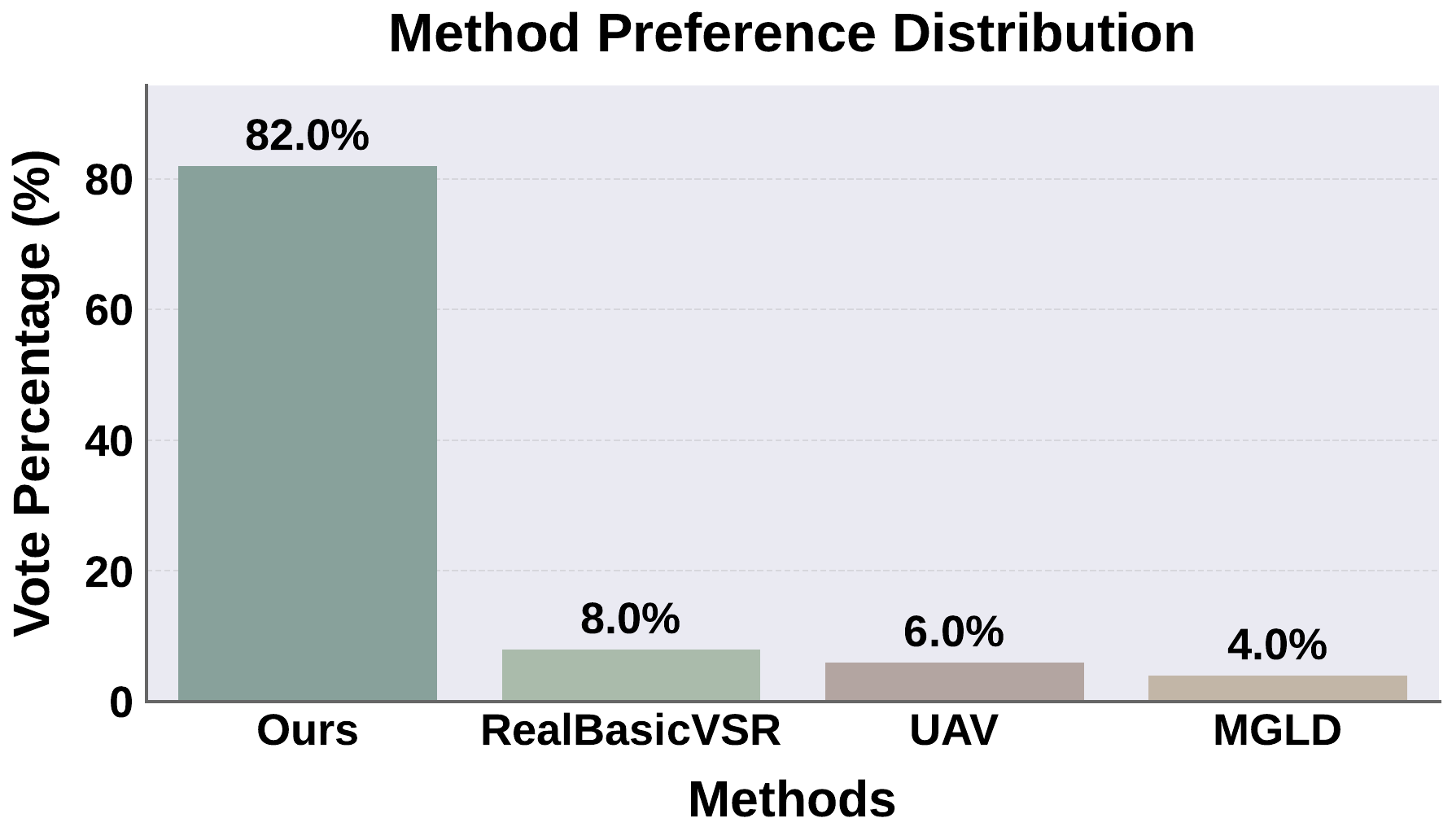}

   \caption{Quantitative comparison of user preferences among different video restoration methods. Our method achieves the highest preference rate (82.0\%) in the user study, significantly outperforming RealBasicVSR~\cite{chan2022investigating}, MGLD~\cite{yang2025motion}, and UAV \cite{zhou2024upscale}.}
   \label{fig:vote_distribution}
\end{figure}

\subsection{Effectiveness of Text Prompts}

\begin{figure*}[htbp]
  \centering
   \includegraphics[width=0.95\linewidth]{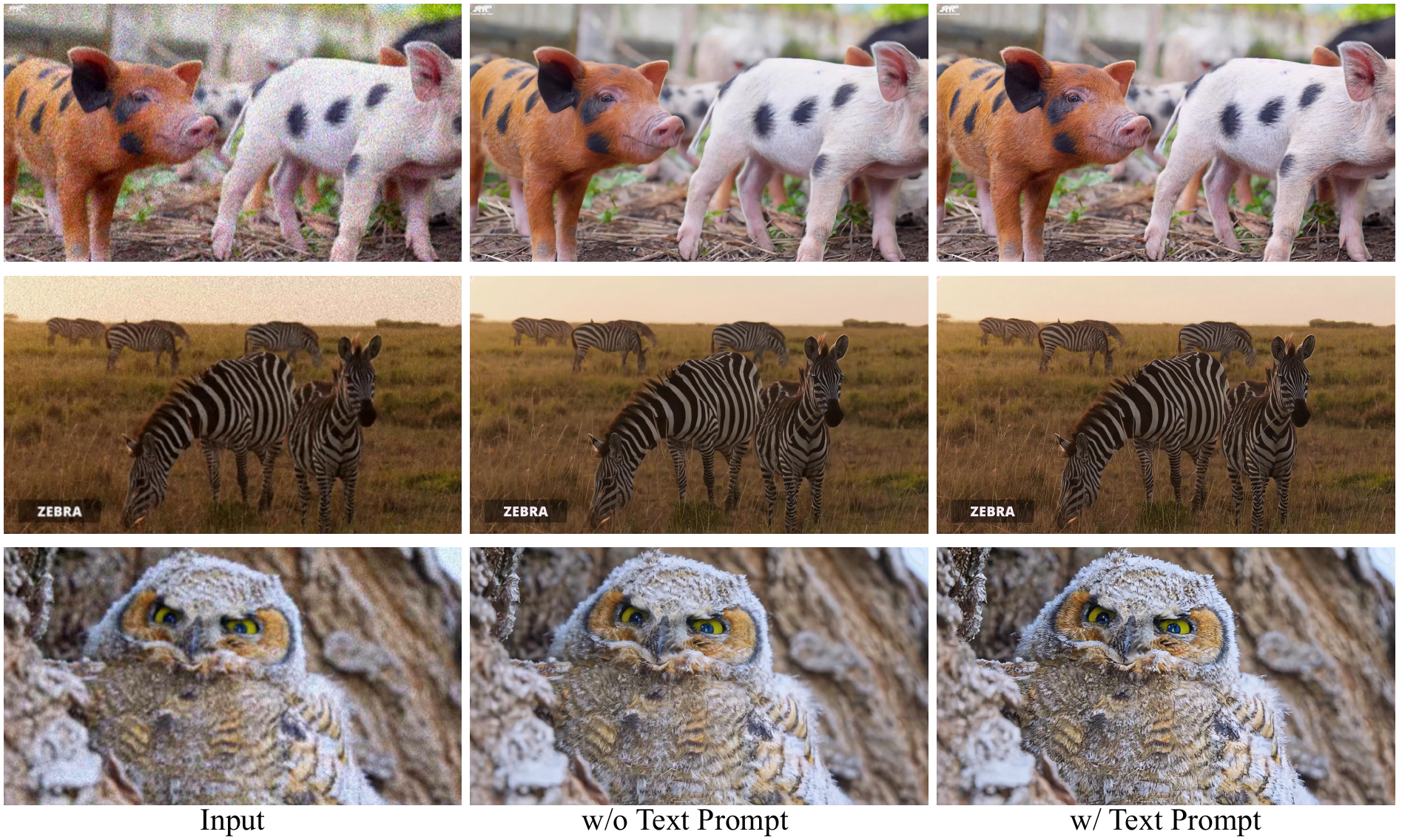}

\caption{Visual comparison of restoration results with and without text prompts. Incorporating meaningful text prompts as guidance significantly enhances visual quality, improving texture sharpness, fine-grained details, and overall realism. \textbf{(Zoom-in for best view)}}
   \label{fig:wowotxt}
\end{figure*}

Our DiffVSR model supports both text-guided and text-free restoration approaches. As shown in Fig.~\ref{fig:wowotxt}, incorporating classifier-free guidance \cite{ho2022classifier} with meaningful text prompts significantly enhances visual quality compared to using empty prompts. This improvement is particularly evident in fine-grained details, such as the enhanced fur textures of the piglet and owl, sharper vegetation in the zebra scene, and more distinct zebra patterns in distant regions. These results demonstrate that appropriate text prompts can effectively guide the restoration process, enhancing intricate textures while maintaining a natural and realistic appearance.

The ability of text prompts to improve results highlights their role in activating the pretrained generative priors within our framework, enabling the model to leverage its full potential when dealing with severely degraded videos. Unlike previous approaches that may struggle with complex degradations regardless of text guidance, our Progressive Learning Strategy creates a foundation where text prompts can more effectively steer the restoration process. These findings align with prior work \cite{zhou2024upscale}, further validating that text guidance combined with an effective learning strategy can significantly boost performance for complex video restoration tasks.

\section{User Study}
We conducted a user study to evaluate perceptual quality through blind comparison. Twenty participants with a computer vision background were asked to compare restoration results from RealBasicVSR \cite{chan2022investigating}, MGLD \cite{yang2025motion}, UAV \cite{zhou2024upscale}, and our DiffVSR. The study included 20 test sets, containing both real-world and synthetic degraded videos. Participants were instructed to select the most visually appealing result based on three criteria: overall fidelity, detail preservation, and temporal consistency. To ensure unbiased evaluations, method names were hidden, and display order was randomized. As shown in Fig.~\ref{fig:vote_distribution}, our method achieved the highest preference rate across various degradation scenarios, particularly on severely degraded videos where competing methods struggle. This validates our approach of addressing the fundamental learning burden through Progressive Learning Strategy rather than solely focusing on architectural complexity.

\begin{table*}[ht]
\centering
\resizebox{0.95\textwidth}{!}{%
\begin{tabular}{c|c|cccccccc}
\toprule
Datasets & Metrics & Real-ESRGAN\cite{wang2021real} & SD ×4 Upscaler\cite{rombach2022stable} & DiffBIR\cite{lin2023diffbir} & RealBasicVSR\cite{chan2022investigating} & MGLD\cite{yang2025motion} & VEnhancer\cite{he2024venhancer} & UAV\cite{zhou2024upscale} & Ours \\
\hline
\multirow{4}{*}{\centering VideoLQ} & NRQM↑ & 5.772 & 4.956 & \textcolor{red}{6.430} & 6.175 & 5.990 & 4.351 & 5.090 & \textcolor{blue}{6.257} \\
& CLIP-IQA↑ & 0.633 & \textcolor{blue}{0.682} & 0.611 & 0.669 & \textcolor{red}{0.698} & 0.659 & 0.639 & \textcolor{blue}{0.682} \\
& MUSIQ↑ & 49.837 & 34.864 & 53.575 & \textcolor{blue}{55.975} & 47.958 & 43.555 & 36.369 & \textcolor{red}{57.812} \\
& DOVER↑ & 0.728 & 0.518 & 0.679 & \textcolor{blue}{0.742} & 0.730 & 0.657 & 0.632 & \textcolor{red}{0.747} \\
\bottomrule
\end{tabular}%
}
\caption{Quantitative comparison with state-of-the-art methods on VideoLQ dataset.}
\label{tab:quality_videolq}
\end{table*}

\section{Trade-off Between Visual Quality and Temporal Consistency}
In video restoration tasks, achieving a balance between visual quality and temporal consistency is a well-known challenge~\cite{zhou2024upscale,yang2025motion,liu2024temporally}, akin to the perception-distortion trade-off in image restoration~\cite{Blau_2018_CVPR}. Higher restoration quality typically enhances texture detail but can often lead to reduced temporal consistency across frames.

\begin{figure}[htbp]
  \centering
   \includegraphics[width=\linewidth]{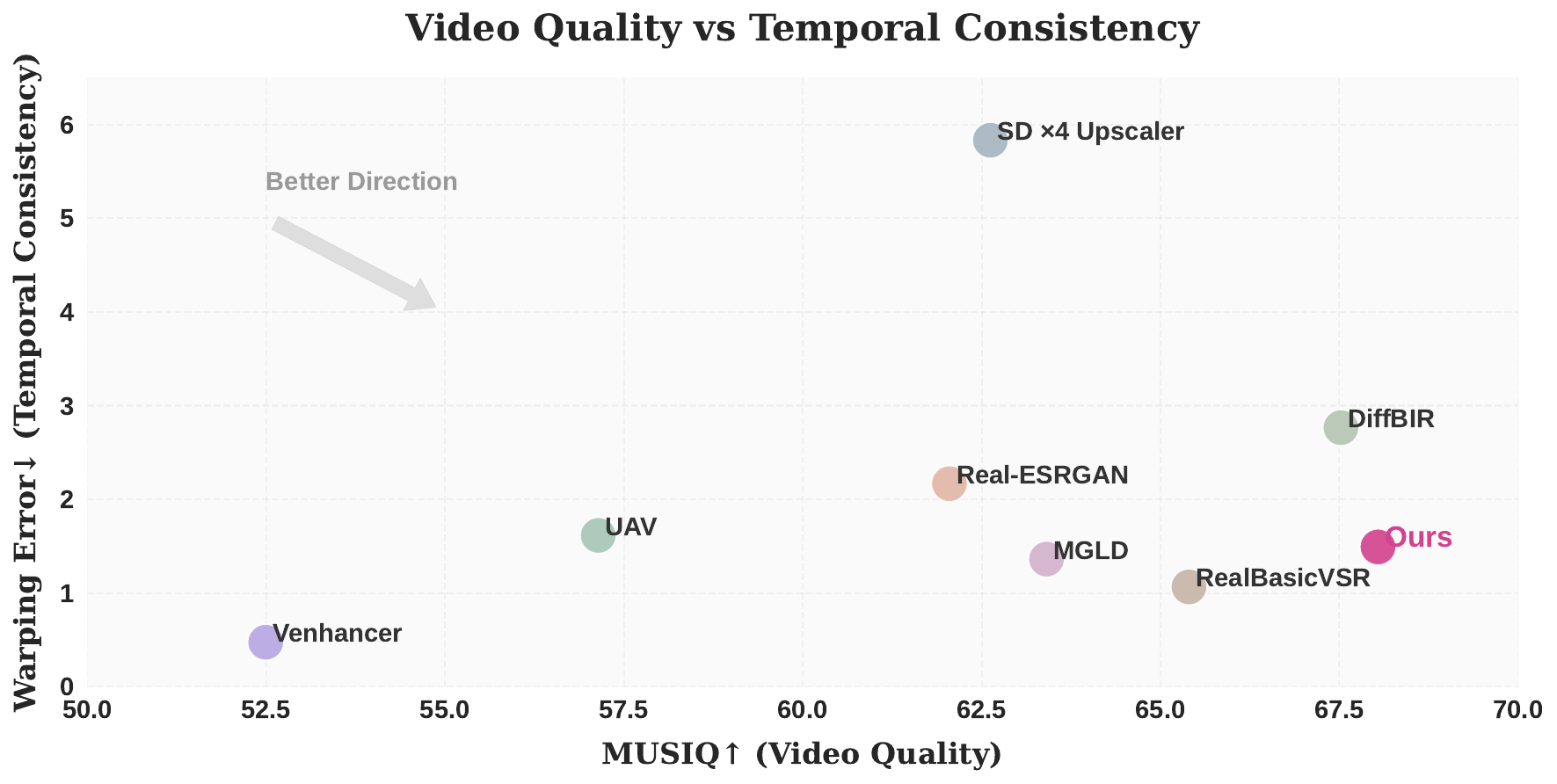}

    \caption{Analysis of the trade-off between video quality (MUSIQ) and temporal consistency (Warping Error) on YouHQ40 dataset. Methods towards the lower-right corner achieve better overall performance, with higher MUSIQ and lower warping error.}
   \label{fig:tradeoff}
\end{figure}

To illustrate this trade-off, we analyze MUSIQ (video quality) and Warping Error (temporal consistency) scores on the YouHQ40 dataset, as shown in Fig.~\ref{fig:tradeoff}. The ideal performance lies in the lower-right region, reflecting high video quality and low warping error. As observed, image-based methods achieve higher quality scores but poor temporal consistency, while video-based approaches maintain better temporal coherence at the cost of reduced quality. Our DiffVSR achieves a favorable balance between these competing objectives, demonstrating how our Interweaved Latent Transition technique effectively maintains temporal consistency without additional training overhead. This further supports our thesis that addressing learning strategy and making intelligent design choices can lead to superior performance in complex degradation scenarios, without requiring overly complex architectural modifications.

\section{More Quantitative and Qualitative Comparisons}
\subsection{More Quantitative Evaluation}
We provide additional quantitative evaluations on the VideoLQ~\cite{chan2022investigating} dataset, which contains real-world low-quality videos with diverse degradation types. As shown in Tab.~\ref{tab:quality_videolq}, our DiffVSR consistently outperforms existing approaches across all metrics. Notably, we achieve significant improvements in perceptual quality metrics, with the highest scores in MUSIQ (57.812). The superior DOVER score (0.747) further validates our approach's effectiveness in maintaining both visual quality and temporal consistency. These results align with our findings in the main paper, demonstrating that addressing the learning burden through our Progressive Learning Strategy leads to robust performance across different real-world degradation scenarios.

\subsection{More Qualitative Results}
\begin{figure*}[!h]
  \centering
   \includegraphics[width=0.85\linewidth]{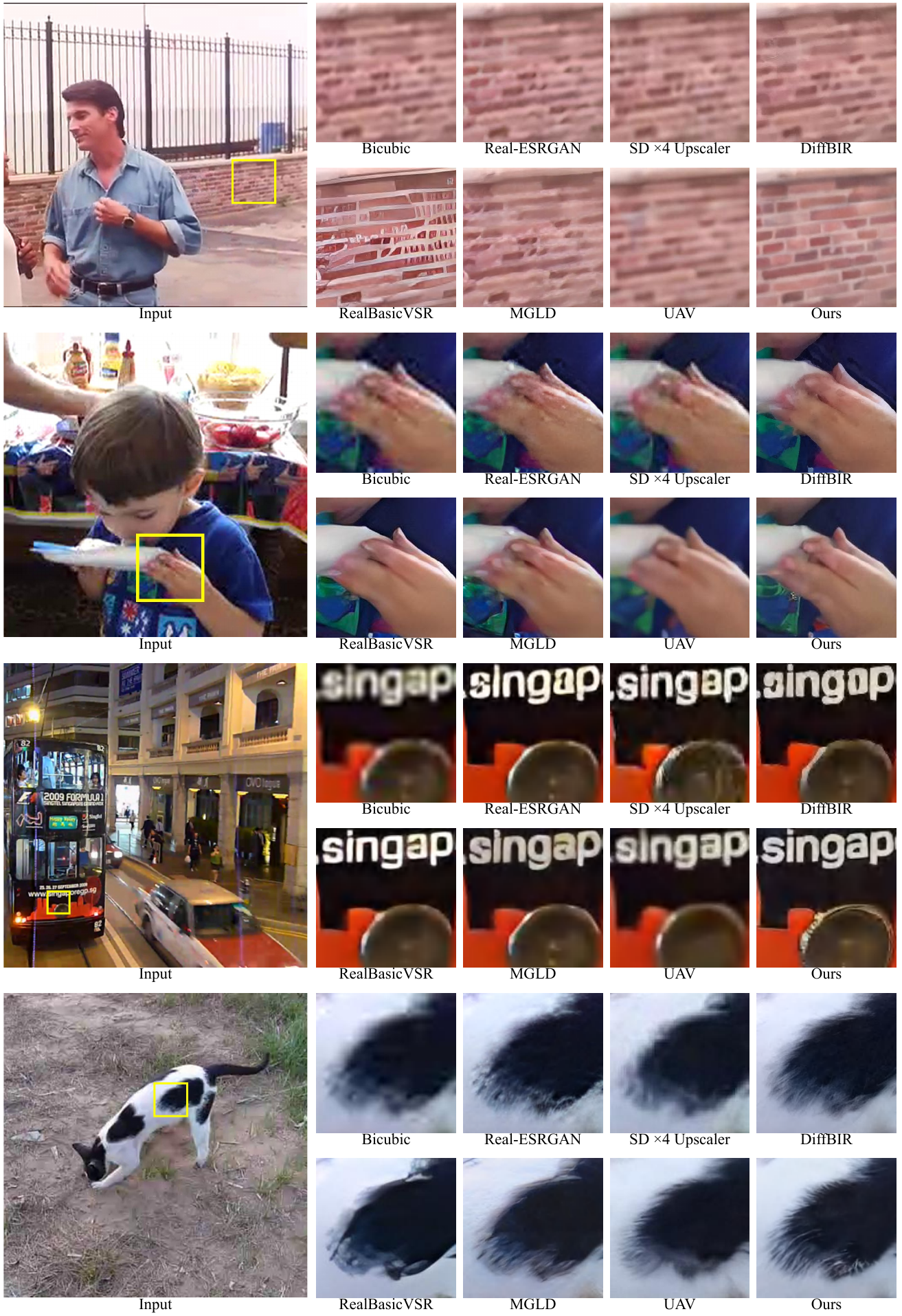}

   \caption{Qualitative comparisons on real-world videos. Our method effectively recovers fine details while maintaining natural textures. \textbf{(Zoom-in for best view)}}
   \label{fig:realcomp}
\end{figure*}

We conduct comprehensive visual comparisons against state-of-the-art approaches, including both image-based methods (ESRGAN~\cite{wang2021real}, SD ×4 Upscaler~\cite{rombach2022stable}, DiffBIR~\cite{lin2023diffbir}) and video-based methods (RealBasicVSR~\cite{chan2022investigating}, MGLD~\cite{yang2025motion}, VEnhancer~\cite{he2024venhancer}, UAV~\cite{zhou2024upscale}). Figs.\ref{fig:syscomp} and\ref{fig:realcomp} showcase the qualitative results on synthetic and real-world test videos, respectively. Our method demonstrates superior capability in recovering diverse textures and structures across various severely degraded scenarios, including architectural elements (wall textures, brick patterns), organic details (facial features, hair strands), natural scenes (vegetation, marine life), and high-frequency patterns (text, dental structures). The restored results exhibit both high fidelity to reference images and rich textural details, while avoiding common artifacts like over-smoothing or false pattern generation that often plague methods struggling with complex degradation modeling.

\section{Video Demo}
We also provide a demonstration video \textcolor{red}{[DiffVSR.mp4]} to showcase the capabilities of our method on both synthetic and real-world videos. The video highlights temporal coherence and dynamic detail preservation enabled by our Interweaved Latent Transition technique, which are better appreciated in motion than through static comparisons. \textbf{The demonstration video is included in the supplementary materials. Note that due to file size limitations, the video has been compressed; the original results exhibit even higher visual quality.}

\section{Limitations}
While DiffVSR achieves significant improvements over existing methods, it has several limitations: (1) As a diffusion-based model, DiffVSR requires repetitive iterations for inference, resulting in slower processing times. This makes it challenging to deploy in real-time applications, though our ILT approach helps maintain competitive computational efficiency compared to alternative temporal consistency methods. (2) DiffVSR struggles with certain challenging scenarios, such as small faces, small human bodies, and complex street scenes, due to inherited limitations of current diffusion-based generative models. Addressing these issues may require further refinements to our Progressive Learning Strategy and additional task-specific adaptations. (3) Due to limited computational resources, we have not conducted larger-scale experiments with increased input sizes or batch sizes. Scaling up the training process with more GPUs could potentially further improve our model's ability to handle even more complex degradation distributions through our staged learning approach.

\begin{figure*}[!h]
  \centering
   \includegraphics[width=0.99\linewidth]{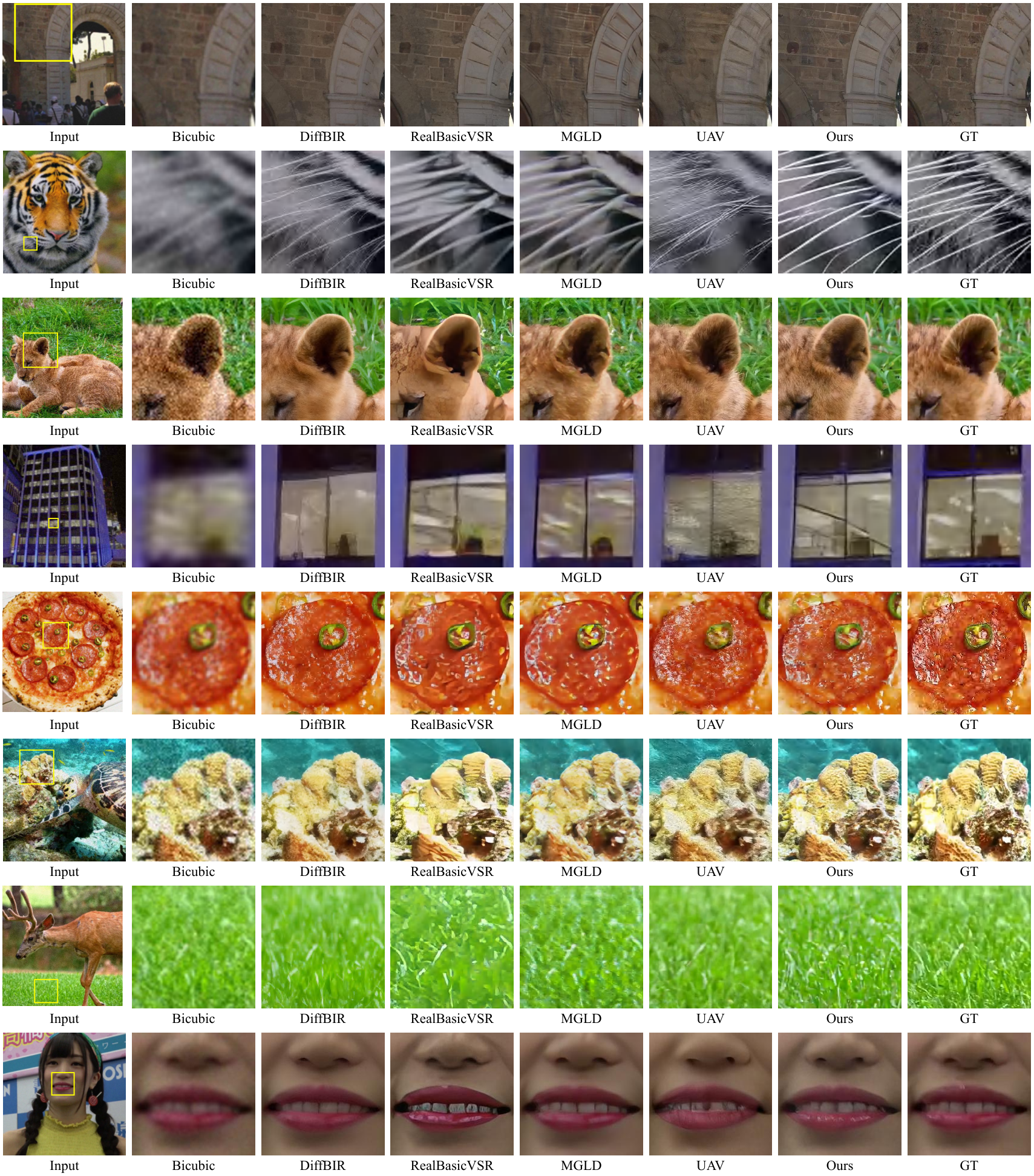}

  \caption{Qualitative comparisons on synthetic datasets. Our method demonstrates superior capability in recovering accurate facial details and textual information, while other methods struggle with either over-smoothing or detail distortion. \textbf{(Zoom-in for best view)}}
   \label{fig:syscomp}
\end{figure*}

\end{document}